# Modeling Events as Machines

Sabah Al-Fedaghi
Computer Engineering Department
Kuwait University
Kuwait
sabah.alfedaghi@ku.edu.kw

*Abstract*—The notion of *events* has occupied a central role in modeling and has an influence in computer science and philosophy. Recent developments in diagrammatic modeling have made it possible to examine *conceptual representation* of events. This paper explores some aspects of the notion of events that are produced by applying a new diagrammatic methodology with a focus on the interaction of events with such concepts as time and space, objects. The proposed description applies to abstract machines where events form the dynamic phases of a system. The results of this nontechnical research can be utilized in many fields where the notion of an event is typically used in interdisciplinary application

*Keywords-Event diagrammatic representation; conceptual modeling; flow-based description*

## INTRODUCTION

The notion of *events* has occupied a central role in modeling and has an influence in computer science and philosophy; therefore, considerable gain can be achieved by further examining this notion.

An event plays a prominent role in various areas of philosophy, from metaphysics to the philosophy of action and mind, as well as in such diverse disciplines as linguistics, literary theory, probability theory, artificial intelligence, physics, and—of course—history. This plethora of concerns and applications is indicative of the prima facie centrality of the notion of an event in our conceptual scheme. [1]

According to [2], Whitehead [3] affirmed that everything is an event. The world is made of events and nothing but events. Even a solid object is an event; or, better, a multiplicity and a series of events.

There has been a continuous debate concerning how events should be conceptualized [4]. Common sense says events are concrete, nonrepeatable entities with a specific location and duration [1]. Some theories treat events as properties or tropes, whereas others view events as special theoretical entities [1]. An event can also be viewed as a theoretical posit in which entities are not explicitly represented or visible in the involved representation [1].

Recent developments in diagrammatic conceptual modeling have made it possible to explore the *conceptual representation* of an event. This paper examines some aspects of the notion of events that result from a new methodology of *diagrammatic representation*. Specifically, the emphasis in the proposed representation of events is on its interaction with the concepts of time and space, objects.

The results of this nontechnical research could be utilized in many fields. The notion of an event is typically used in scientific practice for its latitude, which allows for interdisciplinary circulation and theoretical track-keeping (These terms are from [1]).

In some broad sense we do expect results of research about events … to be at least partially commensurable across disciplines and across time, and this is why researchers tend to go along with umbrella notions rather than more technically refined concepts. [1]

The proposed method of event representation will be based on a diagrammatic model, called the flowthing machine (FM) model, which will be reviewed briefly in the next section. The FM has been utilized as a modeling tool in several fields, such as software engineering, business processes, and engineering design [5-8].

## FLOWTHINGS MACHINE (FM)

The FM model is a diagrammatic schema that uses *flowthings* (hereafter, *things*), defined as *what can be created, released, transferred, processed,* and *received* by means of stages of a flow machine (see Fig. 1). The machine is a generalization of the typical input-process-output model used in many scientific fields (Fig. 2). Things flow through the stages of the machine when they are created by the machine or imported from other machines. Each type of thing has its own machine. For example, *events* are *things* (i.e., that are being created or processed) that flow in their machines, which include submachines such as time and things that are being "eventulized," i.e., going through events.

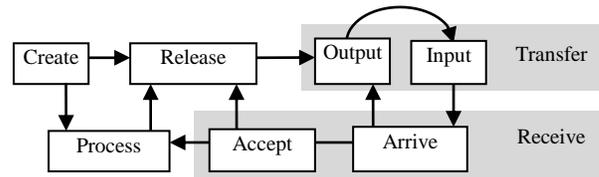

Figure 1. Flow machine

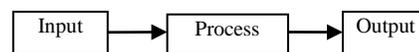

Figure 2. Input-process-output model

The machine is the conceptual fiber used to handle (change though stages) flowthings from their inception or arrival to their de-creation or transmission). Machines form the organizational structure (environment) of whatever is described; especially, in our study, events, time, objects. The machines are embedded in a network of assemblies called *spheres* in which the processes of the machines take place.

The stages in Fig. 1 can be described as follows:
**Arrival**: A thing reaches a new machine.
**Acceptance**: A thing is approved to enter a machine. If arriving things are always accepted, *Arrived* and *Accepted* can be combined as a **Reception** stage.
**Processing** (changed): The thing goes through some kind of transformation that changes it without creating a new thing.
**Release**: A thing is marked as ready to be transferred outside the machine.
**Transfer**: The thing is transported somewhere from/to outside the machine.
**Creation**: A new thing is born (created) in a machine.

In general, a flow machine is thought to be an abstract machine that receives, processes, creates, releases, and transfers things. The stages are mutually exclusive. An additional stage of *Storage* can also be added to any machine to represent the storage of things; however, storage is not an exclusive stage because there can be *stored processed* flowthings, *stored created* flowthings, etc.

Flow machines also use the notions of *spheres and subspheres*. These are the network environments and relationships of machines and submachines. Multiple machines can exist in a sphere if needed. A sphere can be a person, an organ, an entity (e.g., a company, a customer), a location (a laboratory, a waiting room), a communication medium (a channel, a wire). A machine is a subsphere that embodies the flow; it itself has no subspheres.

FM also utilizes the notion of *triggering*. Triggering is the activation of a flow, denoted in the machine diagrams by a *dashed arrow*. It is a dependency relationship among flows and parts of flows. A flow is said to be triggered if it is created or activated by another flow (e.g., a flow of electricity triggers a flow of heat), or activated by another point in the flow. Triggering can also be used to initiate events such as starting up a machine (e.g., remote signal to turn on). Multiple machines can interact by triggering events related to other machines in those machines' spheres and stages

**Example**: According to Shaviro [2], for Whitehead [3] the Cleopatra's Needle on the Victoria Embankment in London is a series of events.

Cleopatra's Needle is actively happening. It never remains the same. "A physicist who looks on that part of the life of nature as a dance of electrons, will tell you that daily it has lost some molecules and gained others, and even the plain man can see that it gets dirtier and is occasionally washed" [3]. At every instant, the mere standing-in-place of Cleopatra's Needle is an event: a renewal, a novelty, a fresh creation. [2]

Fig. 3 shows the corresponding structural/static description. The Cleopatra's Needle is created and processed and takes its course as a thing (circle 1 in the figure). Its electronics dance (2) and its molecules come and go (3) in the eyes of the physicist (4). Dirt flows on it and is cleaned off (5) in the eyes of the layman (6).

A dynamical description is shown in Fig. 4 (it embeds Fig. 3). A series of events (1, 7, 8, …) creates and re-creates the Cleopatra's Needle. *Event 1* (yellow box in the online version of the paper) is a *thing* in its machine that is created and processed (1). Its creation triggers (2) the appearance (creation) of the Cleopatra's Needle with its electronics and dirt. The event has its time (3) that triggers (4) the event creation. Time flows to create its slices (5), (6), which triggers events (7), (8), which in turn trigger the re-creation of the Cleopatra's Needle (9), (10), etc.

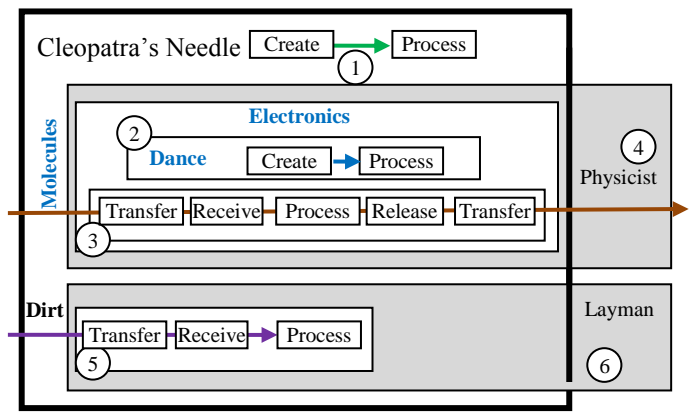

Figure 3. FM representation of *Cleopatra's Needle*.

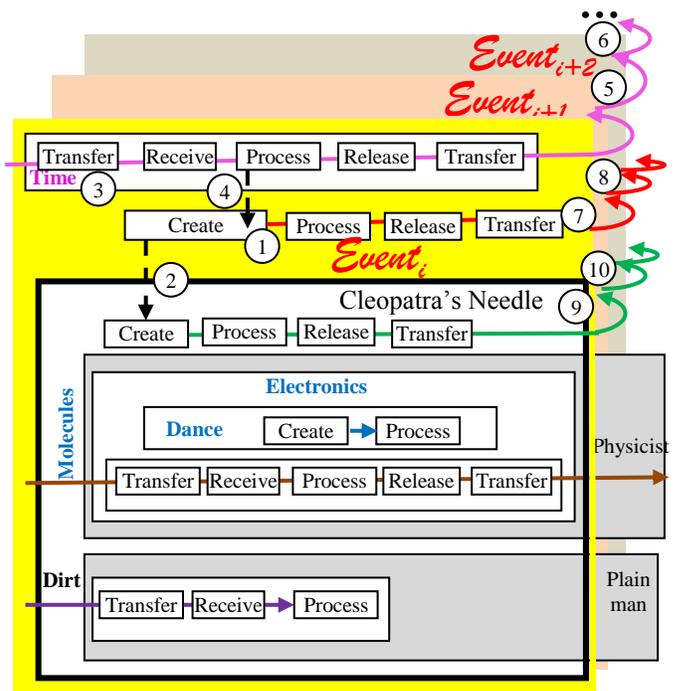

Figure 4. Events of *Cleopatra's Needle*.

## EVENTS AS MACHINES

In the FM, an event is a thing that can be described by a machine such as time flow, past and present, and conceptual spaces.

### A. Defining an event

Consider the ontological theory that "describes both complex spatio-temporal processes and the enduring entities which participate therein" [9]. It comprehends two categories that are comparable to the familiar division in accounting between stocks and *flows* [9].

Consider the sentence *John was in Hyde Park from 6am to 7am*, *on Monday morning*. Fig. 5 shows the FM representation of *John was in Hyde Park*. It is a possible path in reality that describes possible activities. The purpose of such an event-less (no time flow) representation is to bring order to an untidy world and perceive it as consisting of discrete activities and flows that have some orderly relationships.

The *event* that actually occurs for John being in Hyde Park *from 6am to 7am on Monday morning* is a record of what happens. It can be represented as shown in Fig. 6. Its representation is constructed from:
- The event itself (circle 1) as it is created and processed (takes its course)
- The event-related things (e.g., John, Hyde Park) (2 – Fig. 5)
- The time machine of the event (3)

An event in FM is a *thing* with its machine that realizes schemata of activities into a *time being* (happening through time machine) and continues into being (existence) until the end of the event. Of course events can also have *property machines* (e.g., slowness).

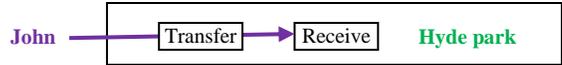

Figure 5. FM representation of *John was in Hyde Park*

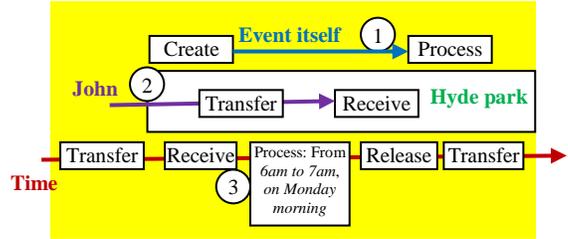

Figure 6. FM representation of *John was in Hyde Park*.

Bittner and Smith [9] refine the level of granularity of this example to the level where the same entity John might be recognized as *John-entering-the-park*, *John-walking-to-his-favorite-bench*, *John-sitting-down-on-his-favorite-bench*, *John-walking-to-the-exit*, and *John-exiting-the-park*. Fig. 7 shows these processes and Fig. 8 shows them as events:
  $V_1$: John-entering-the-park,
  $V_2$: John-walking-to-his-favorite-bench,
  $V_3$: John-sitting-down-on-his-favorite-bench,
  $V_4$: John-walking-to-the-exit, and
  $V_5$: John-exiting-the-park

For simplicity's sake *the stages of the events themselves and time flow are not shown in the figure*.

Accordingly, in the FM approach, there is a dichotomy between the diagram and its events just as there is an association between a class and its objects in the object-oriented paradigm. Event things have their machines, parts, and types. They may have gaps, e.g., interruption. A *trace* is a sequence or group (to include parallelism) of events. It can be considered an event itself.

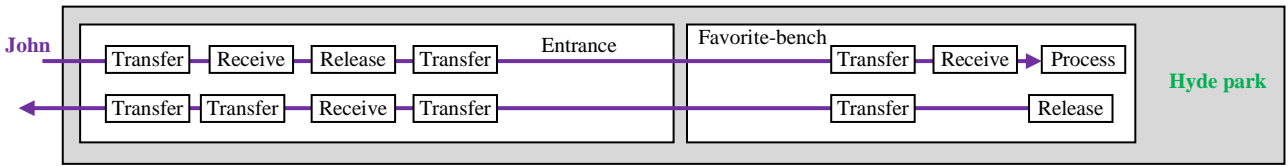

Figure 7. FM representation of *John-entering-the-park, John-walking-to-his-favorite-bench, John-sitting-down-on-his-favorite-bench, John-walking-to-the-exit, and John-exiting-the-park*.

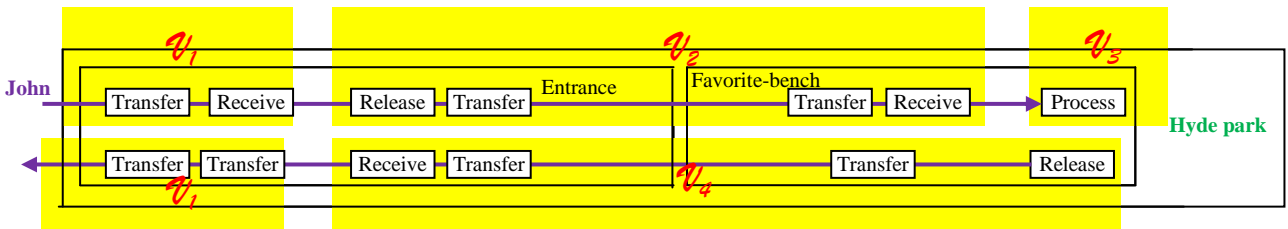

Figure 8. FM representation of the events *John-entering-the-park, John-walking-to-his-favorite-bench, John-sitting-down-on-his-favorite-bench, John-walking-to-the-exit,* and *John-exiting-the-park*.

## B. Past events

Another example that further illustrates the FM approach to events is as follows. According to Zacks and Tversky [4], *Phoebe fed a coelacanth* means that *There exists an event x such that Fed (Phoebe, coelacanth, x)*.

In FM (see Fig. 9), *Phoebe fed a coelacanth* is an event (1) in the past time (2) that flows (3) to a later time (4) to be processed (5 - talked about). Note that it is not necessary to mention past or present time in the diagram because this is implied by the flow of time (2, 4).

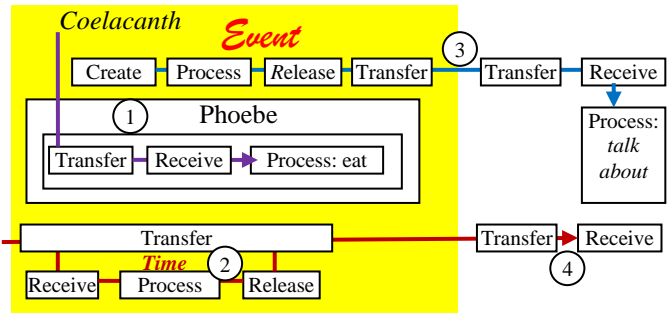

Figure 9. FM representation of *Phoebe fed a coelacanth*.

## C. Events and conceptual space

The notion of *space* in the FM, in general, is viewed as *conceptual* space. Consider the case of a car that is brought to a manufacturing station where two robots process it simultaneously, e.g., one works on the tires and the other on glasses. Fig. 10 shows the FM representation of involved conceptual spheres. The car enters the station (1) and as soon as the car is received there, the car also arrives conceptually, in the spheres of Robot 1 and Robot 2 (2 and 3). Thus, physically, the car is in the sphere of the station, and it is also in the two conceptual spheres of the robots.

Fig. 11 shows possible "meaningful" events (with timing slots) of this example. "Meaningful" here refers to significant events in the modeled context. For example, the event of transferring the car through the door of the station (the *Transfer* stage (1)) seems not be as interesting as the event of *transferring it and receiving it inside the station*. If there is a possibility of the car getting stuck in the door before arrival inside the station, then the mere *transfer* can be considered a separate event.

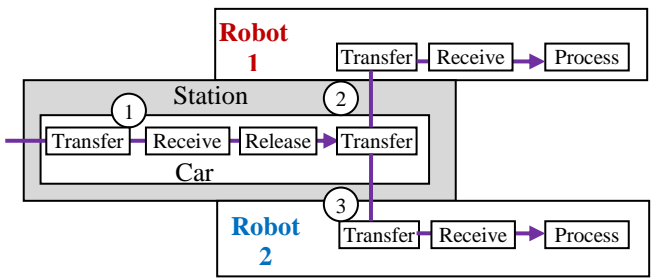

Figure 10. A car can be in two conceptual spaces simultaneously.

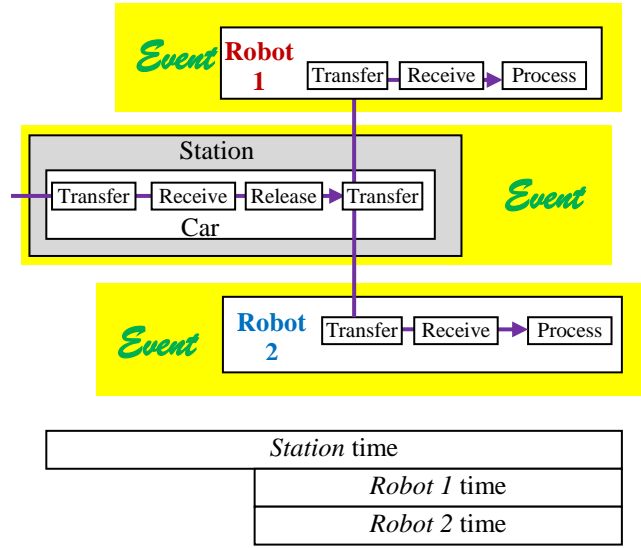

Figure 11. Possible physical and conceptual spaces and times.

This discussion is relevant to Bennett's [10] example as discussed in Zacks and Tversky [4] of the rejection of the "possibility that events might be temporal parts of objects" on the grounds that it fails to cover the case "*if a ball is both heating and rotating, one may wish to refer to these as separate events, though they involve the same thing over the same time-period*" [4].

Fig. 12 shows the FM representation for *A ball is both heating and rotating*. The ball flows in a sphere (e.g., it is thrown in the air by a baseball player), where it receives heat, thus, heating, and rotates. Fig. 13 shows the three events of concern in this context:

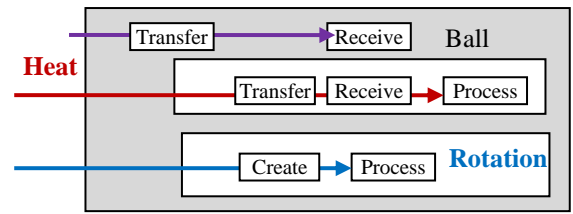

Figure 12. A car in a sphere that is being heated and rotated.

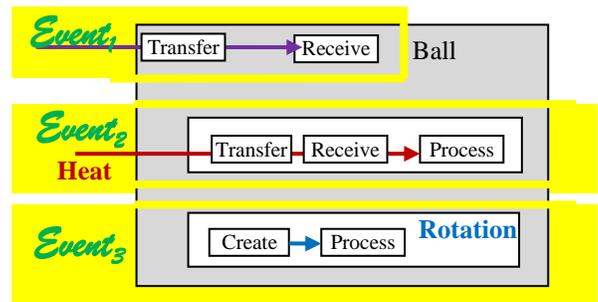

Figure 13. A ball that *is both being heated and rotating*.

- Event₁: The ball is in the sphere, e.g., thrown in the air.
- Event₂: The ball gains heat.
- Event₃: The ball rotates.

It is clear that it is possible to count these as separate events or we may view (Event₂, Event₂) as one combined event, similar to *the car* in the previously discussed example that existed in the separate conceptual spheres of two robots simultaneously.

## EVENTS AND CAUSES

This section explores utilizing the FM in *cause*-related situations. The purpose is not to study the notion of cause, but rather to demonstrate how a different representation exposes different views of causes.

*A. Shankman's example*

Consider the example given by Shankman [11]: "the car's failure to start (an event, it failing) *caused* the professor to arrive late (another event, his arriving)" (Italic added). Fig. 14 shows the corresponding FM representation. First, the professor enters the car (circle 1), then he/she tries to start the car (2). If it fails to start, then this triggers releasing the car to be repaired (3 and 4) and returned (5). If the professor enters his/her car (1) and he/she starts it (6), then he/she drives it to the college (7) to arrive there (8). The time of arrival is either arrival on time or late.

This FM description is a structural/static schemata of all related constructs of *The car's failure to start caused the professor to arrive late*. We can recognize seven "meaningful – with respect to the example" events in the diagram as shown in Fig. 15.

- Event 1 ($V_1$): The professor enters his/her car and tries to start the car.
- Event 2 ($V_2$): The car fails to start and is repaired.
- Event 3 ($V_3$): The car starts.
- Event 4 ($V_4$): The car is driven to the college.
- Event 5 ($V_5$): The car arrives on time and the professor enters the college.
- Event 6 ($V_6$): The car arrives late and the professor enters the college.

Fig. 16 shows the "without delay" trace of events in Fig. 15. *The car's failure to start caused the professor to arrive late* refers to the trace in Fig. 17.

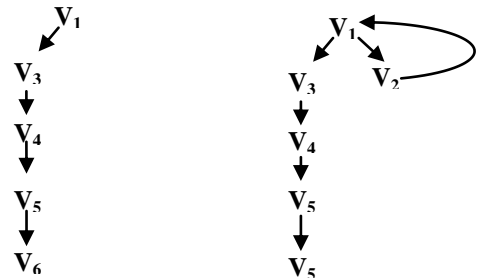

Figure 16. Trace without delay.  Figure 17. Trace with delay.

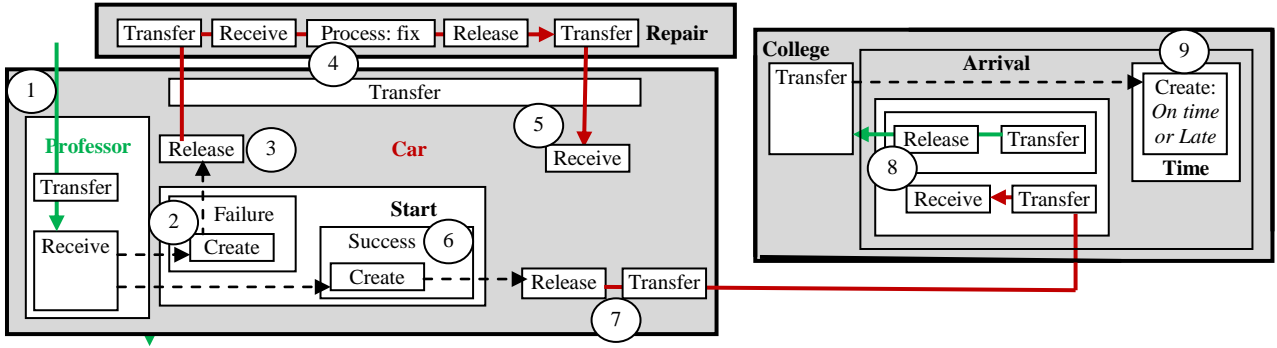

Figure 14. FM representation of the example.

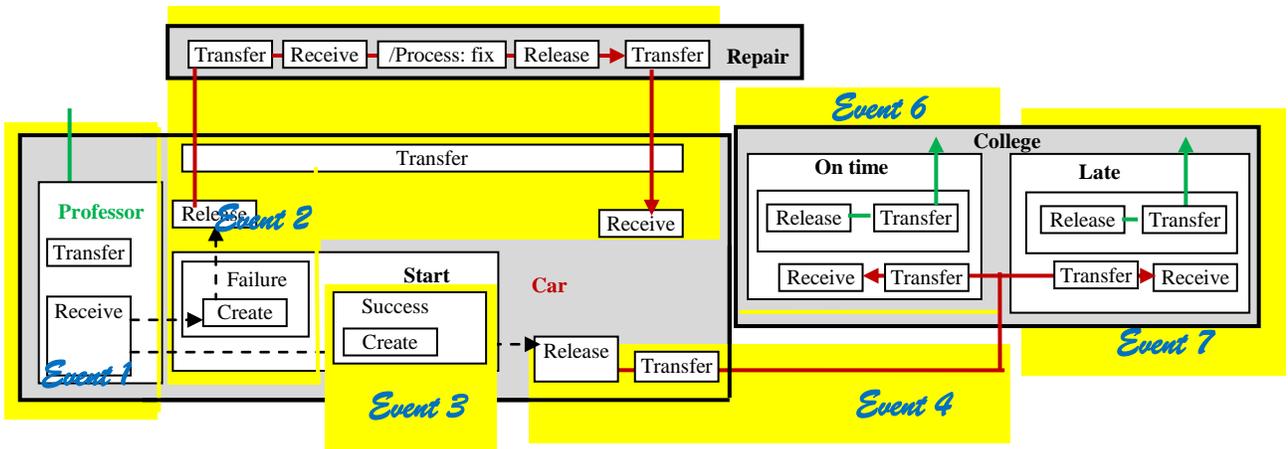

Figure 15. Seven meaningful events in the example.

Note that in this last trace, $V_1$ occurs twice (with two different time stamps). Without a delay ($V_2$), the trace would be,

$$V_1, V_3, V_4, V_5, V_6$$

This trace is a sub-trace of the previous trace.

Consider *The car's failure to start **caused** the professor to arrive late* again as $V_2$ caused $V_6$. From the FM's point of view, it could be explained that *additional events (larger trace length) have increased the trace time*.

*B. Lewis's example*

To further illustrate the significant of the **cause** notion, consider Lewis's [12] example discussed in Zeleňák [13]. It involves the two events *John's walking* and *John's walking slowly*.

> *John's walking slowly* implies *John's walking* because it is necessary that if the first event occurs in some region, then also the second one occurs in the same region [12]. But then it holds that, if the second event had not occurred, the first event would not have occurred either. To avoid the awkward conclusion that the second event caused the first one, Lewis says we may differentiate these two events but we should not take them to be distinct. Since on his counterfactual theory only distinct events stand in causal relations, counterfactual dependence between these two different but not distinct events is noncausal [12]. [13]

From the FM point of view there are a *walk* thing and a *slowness* thing that flow in John. Thus, *John is walking* corresponds to the FM expression *Walk thing flows in John* and *John walking slowly* corresponds to the FM expression *Slowness and walk things flow in John*. *John is walking* is similar to *John is eating*, which indicates that *food flows in John*.

*Walking had not occurred* indicates that John was doing something else *nonwalk* such as standing, lying down, sitting, etc. A walking event will be denoted by the term *walk*, while other events (with the same outfit that goes along with slowness) such as running, jogging, etc. will be denoted by the term *nonwalk*. Similarly, the slowness event will be indicated by the term *slowness*, while other events such as quickness, ordinary pace, etc. will be indicated by *nonslowness*. Accordingly, in the FM representation of Fig. 18 we can recognize the basic events $V_1$, $V_2$, $V_3$, and $V_4$ related to walking and slowness. At the level of an event that is described by two basic events, we have:

Walking slowly ($V_2$, $V_3$)
Walking nonslowly ($V_2$, $V_4$)
Nonwalking slowly ($V_1$, $V_3$)
Nonwalking Nonslowly ($V_1$, $V_4$)

If we focus on *John's walking*, we see that it refers to two diagrams:

*John's walking slowly* ($V_2$, $V_3$) – Fig 19
*Walking nonslowly* ($V_2$, $V_4$) – Fig. 20

Of course, by containment property, *John's walking slowly* "implies" *John's walking*. Additionally, *John's walking* had not occurred, indicates both that *John's walking slowly* had not occurred and *Walking nonslowly* had not occurred.

Lewis [12] mentioned an awkward conclusion:

> If *John's walking* event had not occurred, *John's walking slowly* event would not have occurred either.

This is generated by the possibility that "*John's walking* event not occurring refers to *John's walking nonslowly*; hence, this does not exclude the possibility the event *John's walking slowly*.

Accordingly, we claim that the awkwardness raised is a problem of representation of events. *John's walking slowly* implies *John's walking*; however, *John's walking* not occurringdoes not imply that *John's walking slowly* would not have occurred because *John's walking* may refer to *John's walking nonslowly*.

Figure 18. The FM representation that involves all situations related to walking and slowness.

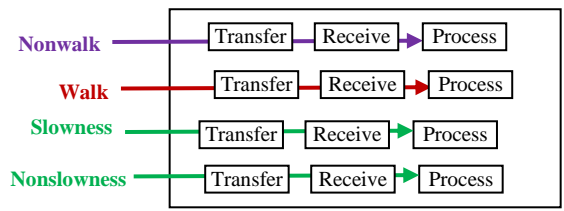

Figure 19. The FM representation of the event *John's walking slowly*.

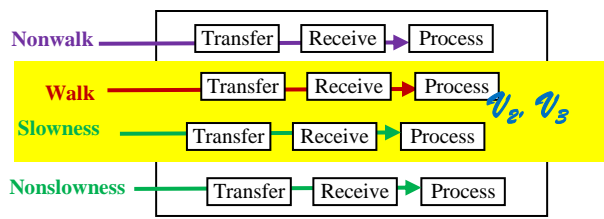

Figure 20. The FM representation of the event *John's walking nonslowly*.

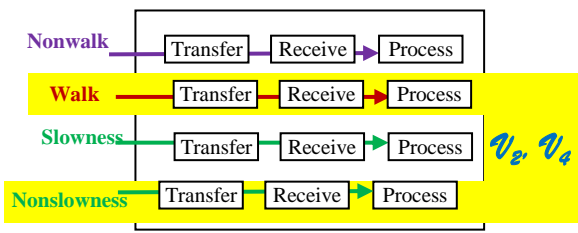

## Conclusion

This paper has explored some aspects of the notion of events by applying a new diagrammatic representation. As a result, new interpretations of cases that involve events have been generated. A diagrammatic definition that encompasses time and things that go through an event was introduced. While the modeled situation of reality produces a static description, events inject the dynamic behavior of the system. Accordingly, with the presentation of the time dimension at this dynamic level, the representation of simultaneous occurrences is possible (e.g., two robots working on the same object simultaneously). Lastly, the proposed representation of events seems to contribute to explaining some cause-related issues, such as the relationship between an event and another event that it contains.

While the paper represents a new direction in alternative representations of event,s we can conclude that the approach is potentially viable to be experienced within the diverse fields that embrace the notion of events. Further research in a specific application of the approach will substantiate our claim of the advantages of the proposed representation.

## References

[1] R. Casati and A. C. Varzi, "Event Concepts," in Understanding Events: From Perception to Action, T. F. Shipley and J. Zacks, Eds. New York: Oxford University Press, 2008, pp. 31–54.

[2] S. Shaviro, "Deleuze's Encounter With Whitehead, The Pinocchio Theory," May 16, 2007. http://www.shaviro.com/Othertexts/DeleuzeWhitehead.pdf

[3] A. N. Whitehead, Adventures of Ideas. New York: Free Press, 1933/1967.

[4] J. M. Zacks and B. Tversky, "Event Structure in Perception and Conception," Psychol Bull, vol. 127, no. 1, pp. 3–21, January 2001.

[5] S. Al-Fedaghi, "Conceptual modeling in simulation: a representation that assimilates events," Int. J. Adv. Comput. Sci. Appl, vol. 7, no. 10, pp. 281–289, 2016.

[6] S. Al-Fedaghi, "Toward a philosophy of data for database systems design," Int. J. Database Theory Appl, vol. 9, no. 10, 2016.

[7] S. Al-Fedaghi, "Diagrammatic modeling language for conceptual design of technical systems: a way to achieve creativity," Int. Rev. Autom. Control, Vol. 9, No. 4, 2016.

[8] S. Al-Fedaghi, "Flowcharting the meaning of logic formulas," *Int.* J. Adv. Res. Artif. Intell. (IJARAI), vol. 5, no. 10, October 2016.

[9] T. Bittner and B. Smith, "Granular Spatio-Temporal Ontologies," American Association for Artificial Intelligence (www.aaai.org), AAAI Technical Report SS-03-03, 2003. http://www.aaai.org/Papers/Symposia/Spring/2003/SS-03-03/SS03-03-003.pdf

[10] J. Bennett, "What events are," in Events, R. Casati and A. C. Varzi, Eds. Aldershot, England: Dartmouth, 1996, pp. 137–151.

[11] P. Shankman, "An event ontology," On Philosophy, February 25, 2009. https://onphilosophy.wordpress.com/2009/02/25/an-event-ontology/

[12] D. Lewis, "Events," in *Philosophical Papers II*, D. Lewis, Ed. Oxford: Oxford University Press, pp. 241–269.

[13] E. Zeleňák, "Two approaches to event ontology," Organon F vol. 16, no. 3, pp. 283–303, 2009.